%% file: main.tex
\documentclass[10pt,twocolumn,letterpaper]{article}
\usepackage{breqn}
\usepackage[pagenumbers]{cvpr} 

\input{preamble}

\definecolor{cvprblue}{rgb}{0.21,0.49,0.74}
\usepackage[pagebackref,breaklinks,colorlinks,allcolors=cvprblue]{hyperref}
\usepackage{url}
\usepackage{amsmath}
\usepackage{graphicx}
\usepackage{array}

\newcommand\blfootnote[1]{%
  \begingroup
  \renewcommand\thefootnote{}\footnote{#1}%
  \addtocounter{footnote}{-1}%
  \endgroup
}

\def\paperID{12268} 
\def\confName{CVPR}
\def\confYear{2025}

\title{PosterMaker: Towards High-Quality Product Poster Generation \\ with Accurate Text Rendering}

\author{
Yifan Gao$^{1,2*\dagger}$,\; Zihang Lin$^{2*}$,\; Chuanbin Liu$^{1\ddagger}$,\; Min Zhou$^{2}$ \\ Tiezheng Ge$^{2}$ ,\; Bo Zheng$^{2}$ ,\;Hongtao Xie$^{1}$ \\
$^{1}$University of Science and Technology of China\; 
$^{2}$Taobao \& Tmall Group of Alibaba \\
\texttt{\small eafn@mail.ustc.edu.cn} \;
\texttt{\small\{liucb92, htxie\}@ustc.edu.cn} \\
\texttt{\small \{linzihang.lzh, yunqi.zm, tiezheng.gtz, bozheng\}@alibaba-inc.com} \\
{\tt\small Project page: \url{https://poster-maker.github.io}}
}

\begin{document}

\twocolumn[{%
	\maketitle
	\renewcommand\twocolumn[1][]{#1}%
	\begin{center}
        \vspace{-1.0 cm}
		\centering        \includegraphics[width=0.94\textwidth]{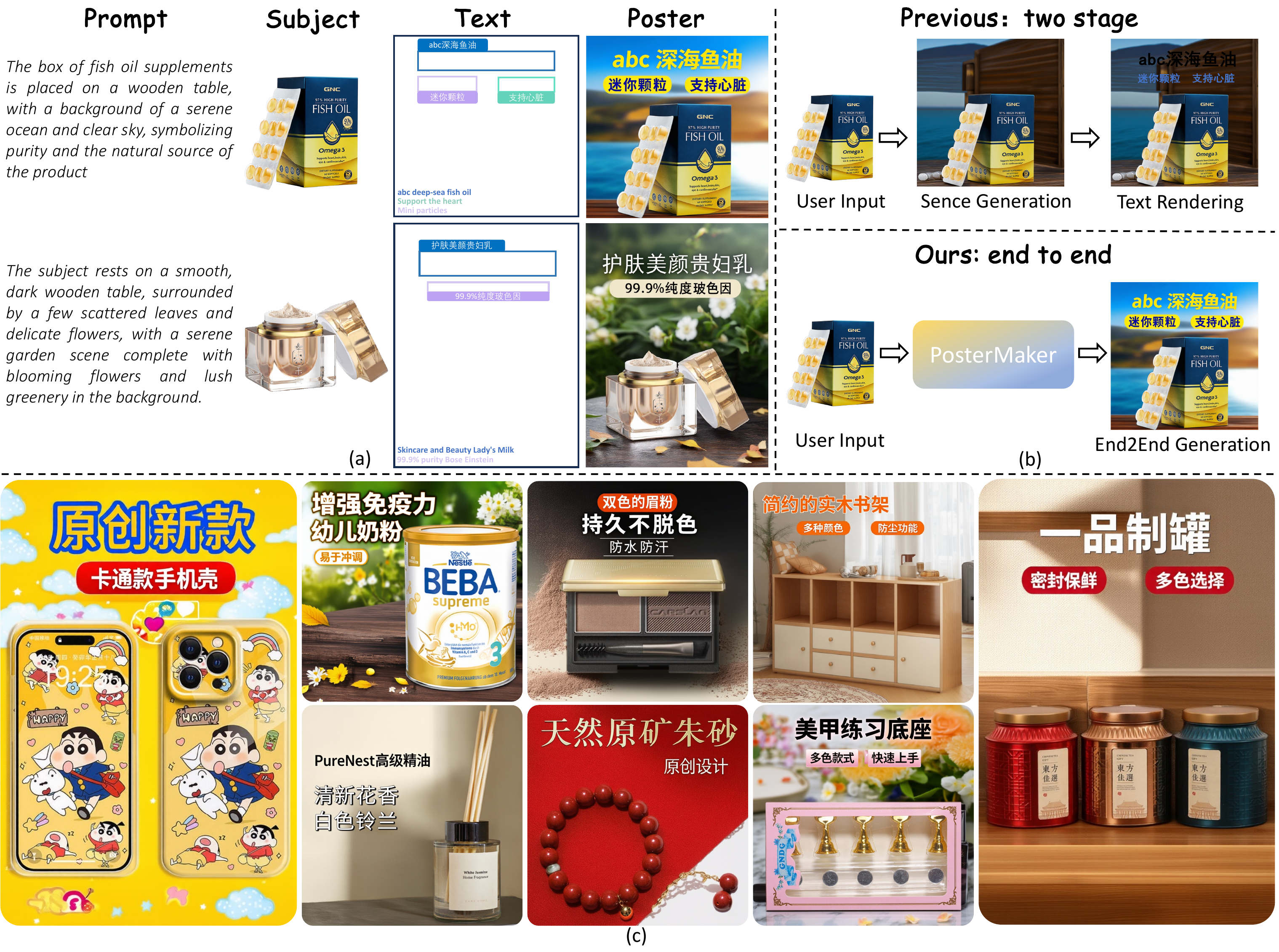}
		 \vspace{-0.5cm}
        \captionof{figure}{(a) Definition of the advertising product poster generation task. The input includes the prompt, subject image, and the texts to be rendered with their layouts. The output is the poster image.
        (b) The comparison of our method with the previous method. PosterMaker generates  posters end-to-end, while previous methods first generate poster backgrounds and then render texts. (c) Visualization results demonstrate that PosterMaker can generate harmonious and aesthetically pleasing posters with accurate texts and maintain subject fidelity.
}
		\label{fig:teaser}
	\end{center}
}]

\blfootnote{$^*$ Equal contribution.\ \ $\ddagger$ Corresponding author.}
\blfootnote{$\dagger$ Work done during the internship at Alibaba Group.}

\input{sec/0_abstract}    
\input{sec/1_intro}

\input{sec/2_related_work}

\input{sec/3_method}
\input{sec/4_experiments}
\input{sec/6_conclusion}
\input{sec/acknowledgments}

{
    \small
    \bibliographystyle{ieeenat_fullname}
    \bibliography{main}
}

\input{sec/X_suppl}

\end{document}

%% file: preamble.tex
\usepackage{lineno}

%% file: sec/0_abstract.tex
\vspace{-5mm}
\begin{abstract}
\vspace{-6mm}

Product posters, which integrate subject, scene, and text, are crucial promotional tools for attracting customers. Creating such posters using modern image generation methods is valuable, while the main challenge lies in accurately rendering text, especially for complex writing systems like Chinese, which contains over 10,000 individual characters. In this work, we identify the key to precise text rendering as constructing a character-discriminative visual feature as a control signal. Based on this insight, we propose a robust character-wise representation as control and we develop TextRenderNet, which achieves a high text rendering accuracy of over 90\%. Another challenge in poster generation is maintaining the fidelity of user-specific products. We address this by introducing SceneGenNet, an inpainting-based model, and propose subject fidelity feedback learning to further enhance fidelity. Based on TextRenderNet and SceneGenNet, we present PosterMaker, an end-to-end generation framework. To optimize PosterMaker efficiently, we implement a two-stage training strategy that decouples text rendering and background generation learning. Experimental results show that PosterMaker outperforms existing baselines by a remarkable margin, which demonstrates its effectiveness. 

\end{abstract}

%% file: sec/1_intro.tex
\vspace{-0.5cm}
\section{Introduction}\label{sec:intro}
Product posters, which showcase items for sale within well-chosen background scenes and include descriptive text, play a vital role in e-commerce advertising by capturing customers' attention and boosting sales. Creating such posters necessitates photographing the product in carefully selected environments that highlight its features, as well as thoughtfully choosing text colors and fonts to ensure that the text is appealing, legible, and harmonious with the background. This process can be quite expensive. With the significant advancements in large-scale text-to-image (T2I) models~\cite{rombach2021highresolution, podell2023sdxl, esser2024scaling}, synthesizing such product posters with image generation models attracts increasing attention. In this paper, we focus on the product poster generation task. Specifically, given a prompt describing the background scene, the foreground image of the user-specified subject and some texts together with their layouts, we aim to develop a model to generate the subject into the desired scene background and accurately render the text in an end-to-end manner (as shown in \cref{fig:teaser} (a)).

\begin{figure*}[!h]
    \centering
  \includegraphics[width=0.9\textwidth]{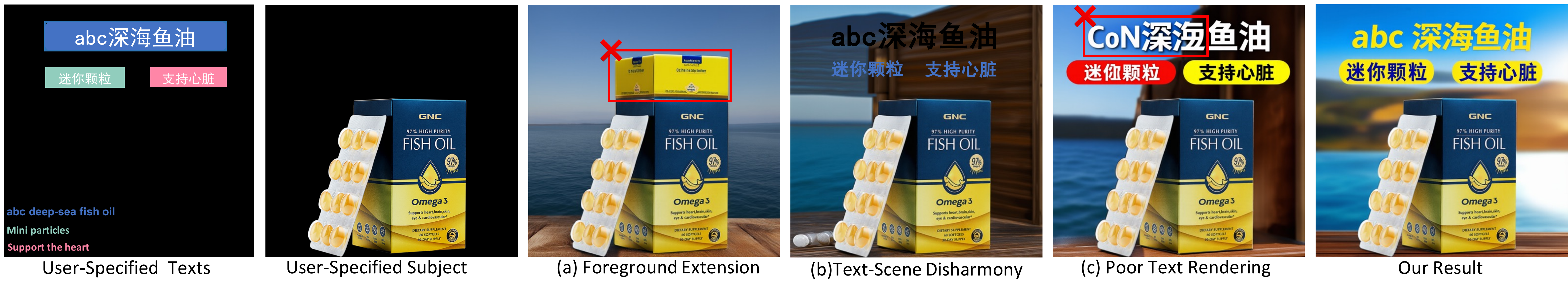}
    \vspace{-0.4cm}
  \caption{The illustration of the three challenges faced by poster generation, which seriously hinder the practical application. }
  \label{fig:intro_challange_show}
  \vspace{-0.5cm}
\end{figure*}

A straightforward solution for this task is to first generate the subject into the desired scene~\cite{du2024towards, DBLP:conf/mm/CaoKZYDZZ24, ruiz2023dreambooth}, and then predict the text attributes (such as color and font)~\cite{lin2023autoposter, gao2023textpainter} and render them on the image. However, such two-stage approach suffers from disharmony between the text and the poster background(as shown in \cref{fig:intro_challange_show} (b)). And collecting training data is also challenging since the text attributes, especially the text font, are difficult to extract from the poster. Another solution is learning to generate the poster using a per-pixel synthesis approach, which can benefit from directly learning the distribution of professionally designed posters. We focus on such one-stage solution. The main challenge is how to ensure the text rendering accuracy.

Many recent works~\cite{esser2024scaling, yang2024glyphcontrol, liu2024glyph, tuo2023anytext} have been proposed to improve the text rendering accuracy for large diffusion models. Great progress has been made and some recent work can achieve high rendering accuracy for English. However, for non-Latin languages like Chinese, one of the most widely spoken languages, achieving high rendering accuracy remains challenging. This difficulty stems from the existence of over 10,000 characters, with Chinese characters characterized by complex and diverse stroke patterns, making it extremely difficult to train a model to memorize the rendering of each individual character. 
Recent studies~\cite{chen2024diffute, tuo2023anytext, ma2024glyphdraw2} have focused on extracting visual features of text as control signals. Typically, these approaches render text lines into glyph images and extract \textbf{line-level} text \textbf{visual} features to guide generation. 

Nevertheless, line-level visual features often lack the discriminative power to capture character-level visual nuances.
To address this limitation, GlyphByT5~\cite{liu2024glyph, liu2024glyph-v2} introduced a box-level contrastive loss with sophisticated glyph augmentation strategies to enhance character-level discriminativeness, achieving promising results. In this paper, we point out that the key to high-accuracy text rendering lies in constructing \textbf{character-discriminative visual features} as control signals. Specifically, we render each character as a glyph image and extract visual features via a visual encoder. These features are then concatenated with positional embeddings to form a character-level representation. Then we propose TextRenderNet, an SD3~\cite{esser2024scaling} controlnet-like~\cite{zhang2023adding} architecture that takes the character-level representation as the control signal to render visual text. Our experiments demonstrate that the proposed character-level representation is effectively capable of achieving accurate text rendering.

  In the task of poster generation, another important thing is to generate the user-specific subject into a desired scene while keeping high subject fidelity. Recent subject-driven controllable generation methods ~\cite{ye2023ip, ruiz2023dreambooth, wang2024instantid} can synthesize such images, but they still cannot ensure that the user-specified subject is completely consistent in the generated details (e.g., the logo on the product may be inaccurately generated), which could potentially mislead customers. Therefore, we follow poster generation methods~\cite{ li2023planning, du2024towards, AnyScene} to address this task via introducing an inpainting-based module named SceneGenNet. However, we found that even using inpainting methods, subject consistency is not always achieved as the inpainting model sometimes extends the subject shape (as shown in~\cref{fig:intro_challange_show} (a)). Similar phenomenon is also observed in ~\cite{Eshratifar_2024_CVPR_outgrowth, du2024towards}. To address this issue, we elaboratively develop a detector to detect the foreground extension cases. Then we employ the detector as a reward model to train the SceneGenNet via feedback learning for further improving subject fidelity.

  Combining the proposed TextRenderNet and SceneGenNet, we develop a framework named PosterMaker that can synthesize the product poster in an end-to-end manner. To efficiently optimize PosterMaker, we introduce a two-stage training strategy to separately train TextRenderNet and SceneGenNet. This training strategy decouples the learning of text rendering and background image generation, thus TextRenderNet and SceneGenNet can focus on their specific tasks. Qualitative results (as shown in \cref{fig:teaser} (c)) demonstrate our training strategy is effective for training PosterMaker and it achieves promising poster generation results.

To summarize, our contributions are as follows:
\begin{itemize}
    \item We proposed a novel framework named PosterMaker, which mainly consists of a TextRenderNet and a SceneGenNet. With a two-stage training strategy, PosterMaker can synthesis aesthetically product posters with texts accurately and harmoniously rendered on it.

    \item We reveal the core of achieving accurate Chinese text rendering is to construct a robust character-level text representation as the control condition. These findings can inspire future research on improving the text rendering abilities of T2I models.

    \item We improve the subject fidelity via subject fidelity feedback learning, which is shown effective in addressing the subject inconsistency issue.
\end{itemize}

%% file: sec/2_related_work.tex
\vspace{-0.2cm}
\section{Related Work}
\subsection{Poster Generation}
Generating posters involves combining various elements like a subject image, a background scene image, and text to ensure the subject and text are prominently and accurately displayed while maintaining an appealing look. Automating this process is quite complex and challenging. Methods like AutoPoster~\cite{lin2023autoposter}, Prompt2Poster~\cite{Prompt2Poster}, and COLE~\cite{cole} break it down into stages: creating images and layout, predicting the visual properties of text, and rendering the poster. These approaches have several steps and often struggle to precisely obtain all the necessary visual attributes like font and color gradients. With the emergence of more advanced generative models~\cite{podell2023sdxl}, methods like JoyType~\cite{li2024joytype}, Glyph-byt5~\cite{liu2024glyph}, and GlyphDraw2~\cite{ma2024glyphdraw2} can directly generate the image and text together at the pixel level based on the poster prompt, text content, and layout. This more streamlined approach can leverage more readily available poster pixel data for training, but there is still room for improvement in terms of the overall poster cohesion and text accuracy. Our method is also a one-stage, direct pixel-level generation approach that simultaneously creates the image and text. However, our focus is on generating posters for a given product subject, where the input includes the subject image, prompt, text content, and layout. In addition to considering text rendering accuracy and overall poster harmony, we also need to maintain the fidelity of the product.

\subsection{Visual Text Rendering}
Recently,  text-to-image~(T2I) models~\cite{saharia2022photorealistic, balaji2022ediff, esser2024scaling} have made significant strides in enhancing English text rendering by introducing stronger text encoders, such as  T5~\cite{T5}. 
However, multilingual text image generation still faces significant challenges due to the large number of non-Latin characters and complex stroke structures. 
Early work ~\cite{yang2024glyphcontrol} has explored the ControlNet-based method~\cite{zhang2023adding}, using low-level visual images such as glyph images as the control signal for text image generation. However, glyph images are easily affected by text size and shape, especially complex stroke details.  Besides, some recent works~\cite{tuo2023anytext,ma2024glyphdraw2, zhang2024control, chen2024diffute, lu2025dhvt, zhu2024visual}  utilize more robust visual features, such as line-level OCR features as control conditions to further improve the text accuracy. 
But the line-level visual features still perform poorly in representing stroke details for each character. To address this issue, GlyphByT5~\cite{liu2024glyph, liu2024glyph-v2} proposes a method with box-level contrastive learning to align the text features extracted from the language model with the features extracted from the visual encoder. To effectively learn such alignment, GlyphByT5 relies on collecting massive amounts of data and developing complex data augmentation strategies for the alignment pre-training, which lacks flexibility. In contrast, in this paper, we reveal that the key to high-accuracy text rendering lies in constructing discriminative character-level visual features. Thus we propose a plug-and-play and robust character-level text representation derived from off-the-shelf OCR encoders, which can accurately represent the visual structure of the text without additional training and enable precise text rendering. 

\begin{figure*}[!h]
  \centering
\includegraphics[width=\textwidth]{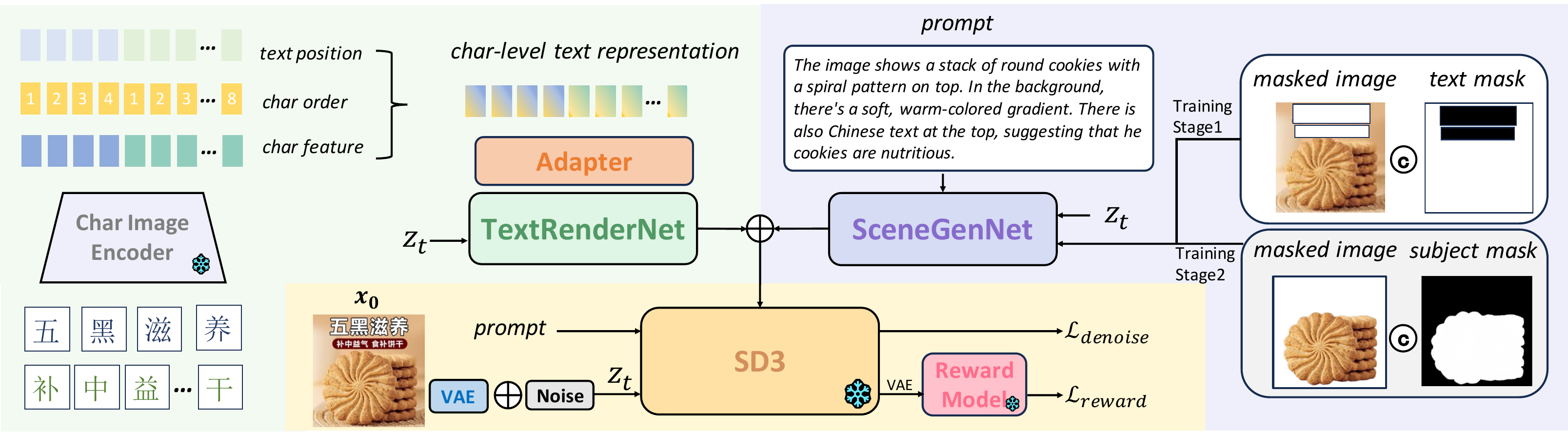}
    \vspace{-0.7cm}
  \caption{The framework of the PosterMaker, which is based on the SD3. To precisely generate multilingual texts and create aesthetically pleasing poster scenes,  TextRenderNet and SenceGenNet are introduced, whose outputs are used as control conditions added to the SD3.}
  \label{fig:method_main}
    \vspace{-0.5cm}
\end{figure*}

\subsection{Subject-Preserved Scene Generation}
To create a scene image with a product subject while ensuring subject fidelity, two main methods are commonly used. One is the subject-driven method ~\cite{ruiz2023dreambooth, DBLP:conf/nips/ChenHLRJCC23, DBLP:conf/cvpr/ChanZJ0W24, DBLP:conf/nips/Li0H23, qi2024deadiff}, which adjusts the position, angle and lighting of the subject based on the prompt to create a harmonious image. However, it often struggles to preserve the significant features of the subject. The other utilizes inpainting-based background completion techniques~\cite{du2024towards, DBLP:conf/mm/CaoKZYDZZ24, DBLP:journals/corr/abs-2312-13309}. It only generates the non-subject areas of an image and naturally keeps consistency in the original subject area. But it sometimes extends the foreground subject~\cite{Eshratifar_2024_CVPR_outgrowth,du2024towards}, such as adding an extra handle to a cup, which also reduces subject fidelity. To maximize subject fidelity, our method uses background completion and a reward model to determine whether the foreground extension occurred, thereby enhancing subject fidelity.

%% file: sec/3_method.tex
\vspace{-0.2cm}
\section{Method}

\subsection{Problem Formulation}

This paper focuses on the creation of product posters, which typically consist of multiple elements such as text, subjects, and scenes, as illustrated in~\cref{fig:teaser} (a). The central challenge of this task is to generate these elements accurately and harmoniously, offering both research and practical applications. The task is defined as: 
\vspace{-0.05cm}
\begin{equation}
     I_g = f(I_{s},M_{s}, T, P ),
\end{equation}
where  $ I_{g} $  denotes the generated poster image, $ I_{s} $ represents the subject image, and $ M_{s} $  is the subject mask. The variable $T$ signifies the content and the position of text and  $ P $  is the prompt describing the background scene. Subsequent sections will detail the design of PosterMaker, and our proposed solution to this task.

\subsection{Framework}
\begin{figure}[t!]
    \centering
    \includegraphics[width=0.45\textwidth]{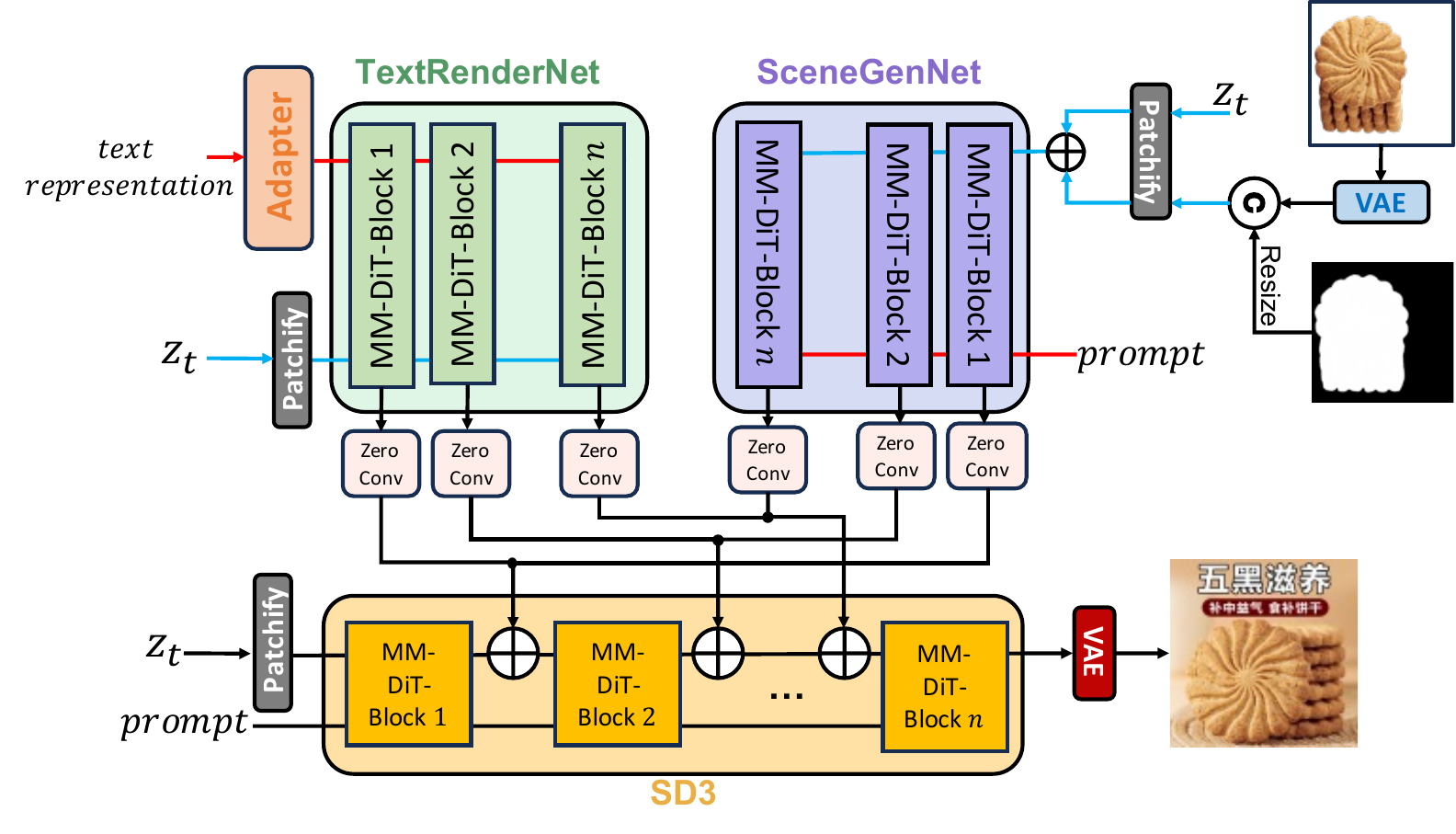}
    \vspace{-0.4cm}
    \caption{The details of TextRenderNet and SceneGenNet, showcasing their model architectures and their interactions with SD3.}
    \label{fig:method_detail}
    \vspace{-0.5cm}
\end{figure}

As shown in~\cref{fig:method_main}, PosterMaker is developed based on Stable Diffusion 3 (SD3)~\cite{esser2024scaling}, which contains a strong VAE for reconstructing the image details like text stroke. And we propose two modules, i.e., TextRenderNet and SceneGenNet, to address the poster generation task. TextRenderNet is specifically designed to learn visual text rendering, taking character-level visual text representations as input to achieve precise and controllable text rendering. SceneGenNet, on the other hand, accepts a masked image (indicating which content should remain unchanged) and a prompt, learning to generate the foreground subject within the desired scene described by the prompt. Both TextRenderNet and SceneGenNet are grounded in a ControlNet-like~\cite{zhang2024control} architecture derived from SD3 and their architectures are detailed in \cref{fig:method_detail}. They share the same internal structure, comprising several cascaded MM-DiT blocks~\cite{esser2024scaling}, with weights copied from the base model for initialization. The output of each MM-DiT block is added to the corresponding block of the base model after passing through a zero convolution layer~\cite{zhang2023adding}. The key distinction between the two modules lies in their input configurations. SceneGenNet takes the prompt as input to the text condition branch, and for the visual branch, the input is derived by the latent feature at timestep $t$, the subject mask, and the masked latent to preserve the foreground area. In contrast, TextRenderNet receives text representations (detailed in the next section) in the text condition branch for text rendering. An adapter, consisting of a linear layer and layer normalization, adjusts the feature dimensions of these text representations before they are input to TextRenderNet. The outputs of each block in TextRenderNet and SceneGenNet are directly added to the corresponding block outputs of the SD3 base model.

\vspace{-0.1cm}
\subsection{Character-level Visual Representation for Precise Text Rendering}
\vspace{-0.1cm}
\label{sec:text_rendering}

\begin{figure}[t!]
    \centering
    \includegraphics[width=0.48\textwidth]{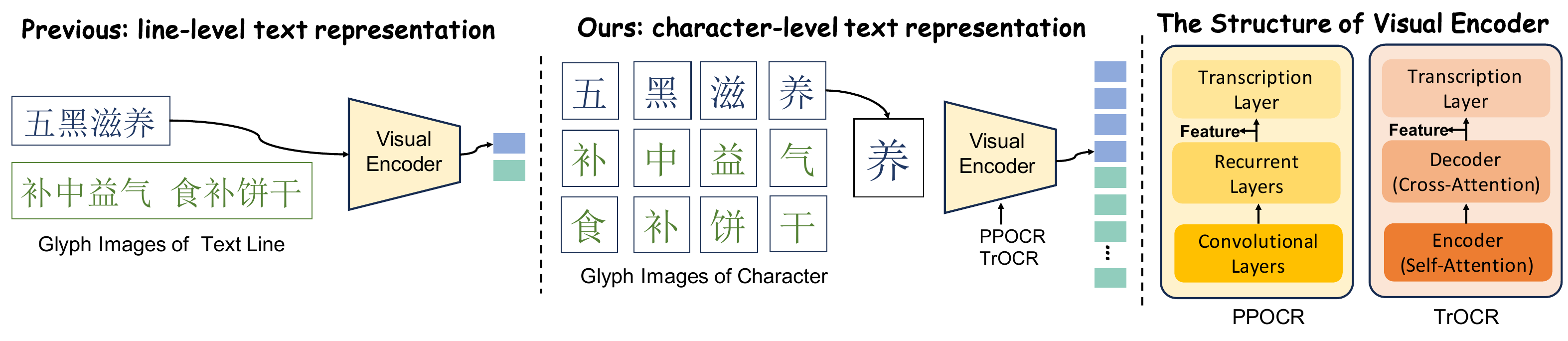}
    \vspace{-0.8cm}
    \caption{The distinction between the multilingual character-level text representation we proposed and the line-level methods of previous works like AnyText~\cite{tuo2023anytext} and GlyphDraw2~\cite{ma2024glyphdraw2}.}
    \label{fig:method_char_encoder}
    \vspace{-0.5cm}
\end{figure}

Recently, some works have explored multilingual visual text generation. Among them, a promising approach is based on ControlNet-like methods~\cite{tuo2023anytext}, which utilize both glyph images and line-level OCR features as conditions. However, this control information cannot accurately represent characters: 1) glyph images are easily affected by text size and shape, making them less robust. 2) line-level visual features  lack fine-grained stroke features and are limited by the OCR model's poor capability to recognize long texts. To address these challenges, this paper proposes a plug-and-play and robust character-level text representation, where each character is precisely represented by one token.

Specifically, the text $C$ has $n$ characters.  For each character $c_i$, its feature is separately extracted by a pre-trained OCR encoder $f_{v}$ and then averaged and pooled to obtain a compact character representation vector  $r_{c_i} \in \mathbb{R}^{c}$. Thus, the character-level text representation is defined as follows:
\vspace{-0.6cm}
\begin{align}
        r_{c_i} = avgpool(f_{v}(I_{c_i})), \\
        R_c = [r_{c_1}, r_{c_2}, ..., r_{c_n}] ,
\vspace{-0.2cm}
\end{align}
where $I_{c_i}$ is the $i$-th character image rendered in a fixed font, and $R_c \in \mathbb{R}^{n \times c}$ is the char-level text representation.

As shown in~\cref{fig:method_char_encoder}, compared to previous methods, our key difference is extracting representations from character glyph images. This enables the model to perceive character stroke structures and achieve high text accuracy. Additionally, since the number of characters is fixed, we can pre-extract the representations of each character and store them in a dictionary, eliminating the need for online rendering and feature extraction. This significantly simplifies the training and inference pipeline.

Finally, this text representation lacks order and positional information. 
Thus, the character order encoding $P_{rank}$ is introduced to represent the order of characters in the text, which is implemented through a sinusoidal position encoding of the char order.  Besides, inspired by GLIGEN~\cite{li2023gligen},  the text position coordinates are mapped to sinusoidal position encoding $ P_{bbox}$ to control the position of the text. Then we concatenate $P_{rank}$, $ P_{bbox}$ and $R_c$ along the feature dimension to construct the final text representation.

\vspace{-0.1cm}
\subsection{Improving Subject Fidelity} \label{sec:subject_faithfulness}
\vspace{-0.1cm}
\begin{figure}[t!]
    \centering
    \includegraphics[width=0.45\textwidth]{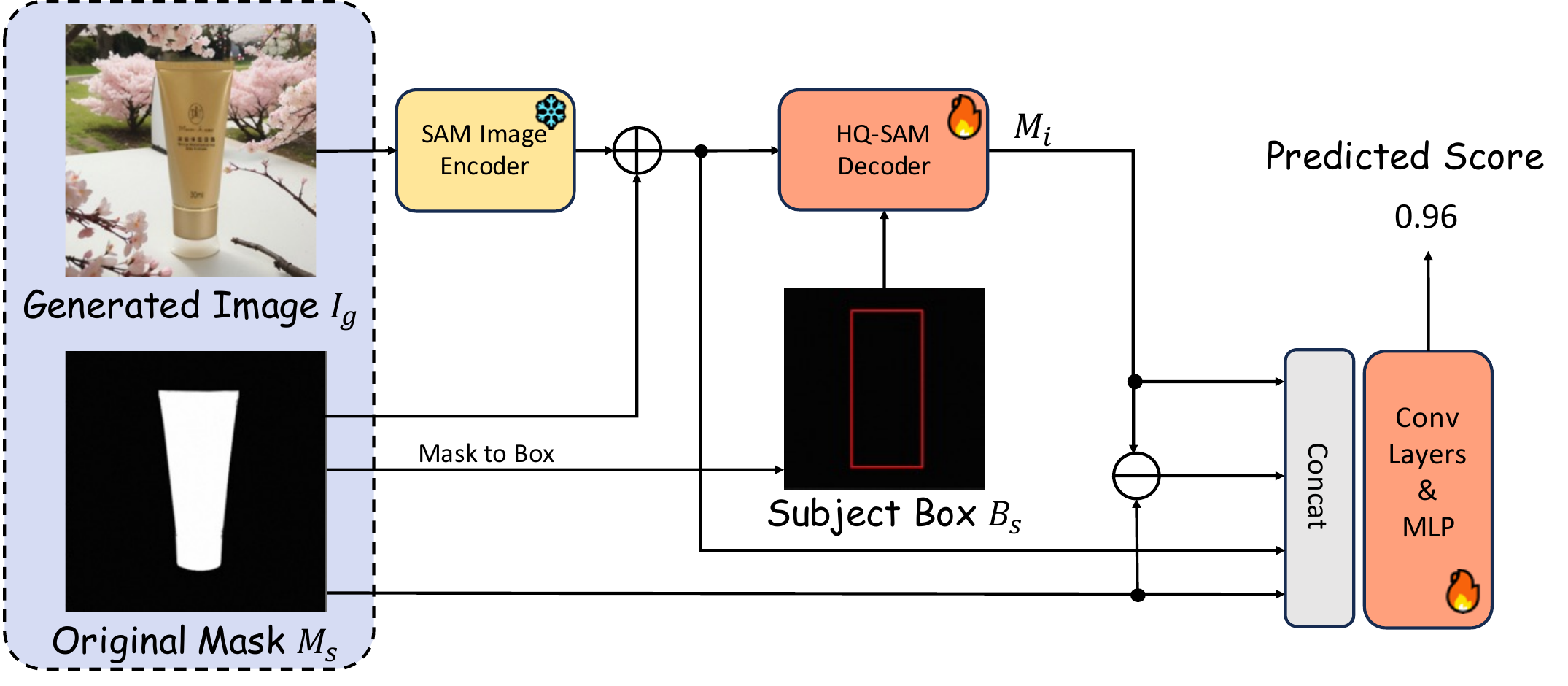}
    \vspace{-0.3cm}
    \caption{
    The model details of the foreground extension detector.}
    \label{fig:reward_model}
    \vspace{-0.3cm}
\end{figure}

\begin{figure}[t!]
    \centering
    \includegraphics[width=1.0\linewidth]{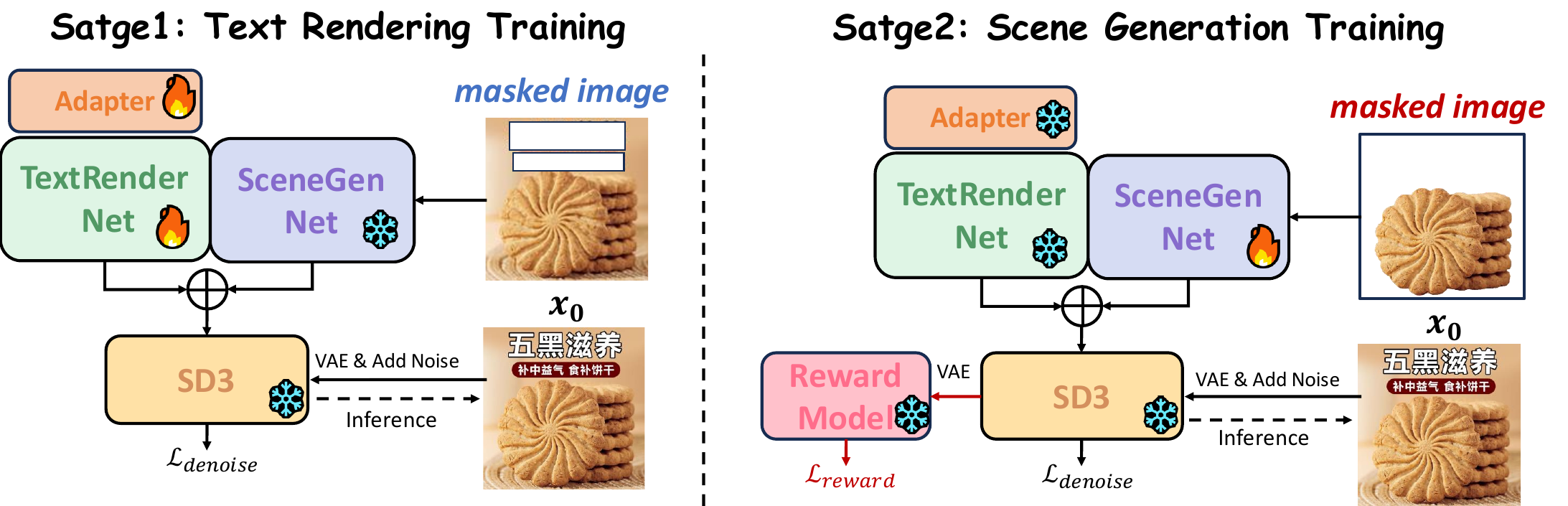}
    \vspace{-0.7cm}
    \caption{
The illustration of our two-stage training strategy for efficiently optimizing PosterMaker.}
    \label{fig:multi_stage_training}
    \vspace{-0.5cm}
\end{figure}

In the task of generating product posters, it is crucial to maintain subject fidelity, i.e., ensuring that the subject in the generated poster remains consistent with the user-specified subject. To achieve this goal, we employ SceneGenNet to perform background inpainting, which is trained to precisely preserve the foreground subject and only inpaint the background according to the prompt. However, inpainting-based models sometimes extend the foreground subject into another subject (as shown in ~\cref{fig:intro_challange_show} (b)) thereby compromising subject fidelity. We refer to this as ``foreground extension''. To mitigate this issue, we develop a model to detect foreground extension and employ it as a reward model to fine-tune PosterMaker to improve subject fidelity.

\noindent \textbf{Foreground Extension Detector.}
We develop the foreground extension detector $S_{\theta}$ based on HQ-SAM~\cite{HQ-SAM}. As shown in \cref{fig:reward_model}, we input the generated image $I_g$ to SAM~\cite{SAM} image encoder. The subject mask $M_s$ and box $B_s$ are provided as mask prompt and box prompt, respectively, to the HQ-SAM decoder to obtain an intermediate mask $M_i$. Next, we concatenate the image features extracted from SAM encoder with $M_s$, $M_i$ and $M_s-M_i$ at the channel dimension. The concatenated features are processed through convolutional layers and MLP layers to predict whether the foreground has been extended in the generated image. We collected 20k manually annotated images to train the foreground extension detector $S_{\theta}$.

\begin{figure*}[!t]
  \centering
\includegraphics[width=\textwidth]{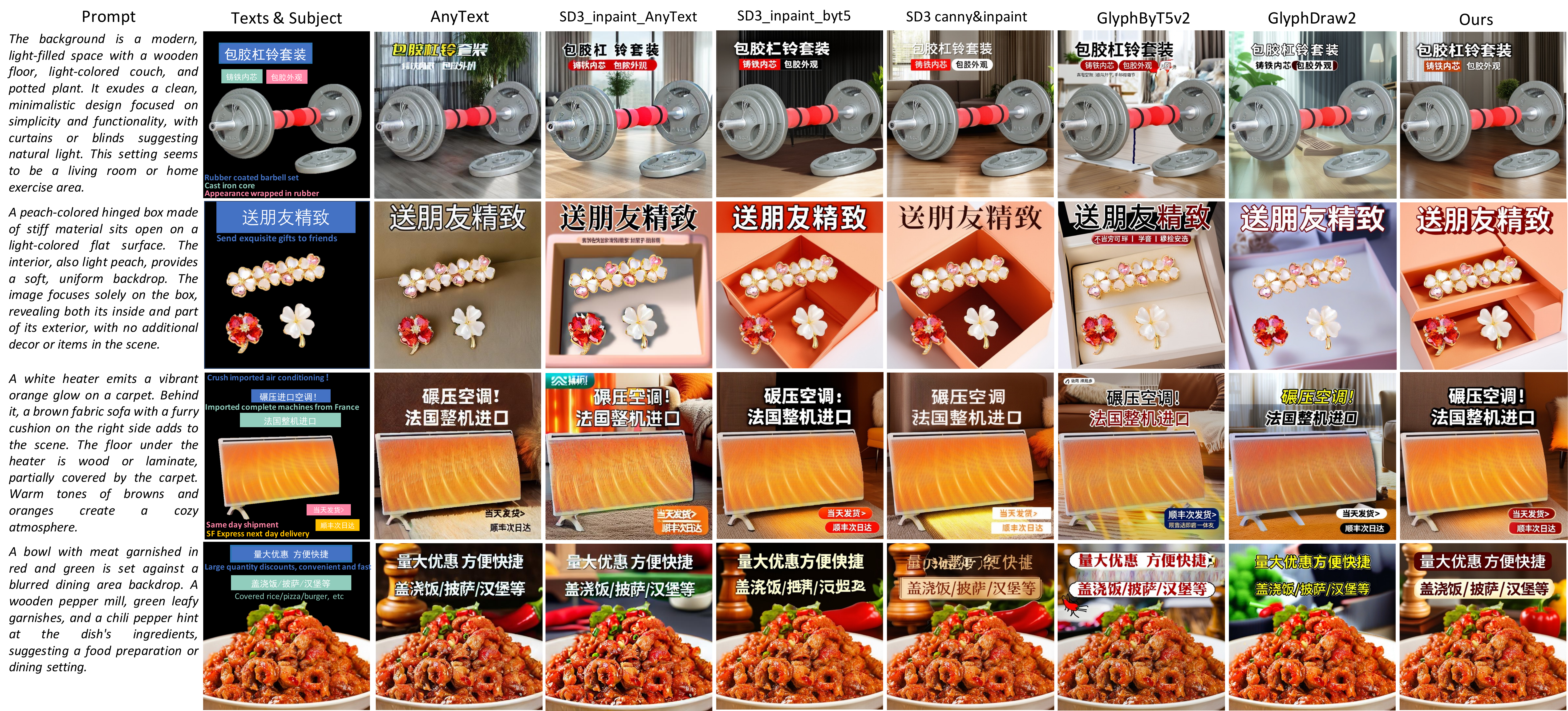}
     \vspace{-0.75cm}
  \caption{Qualitative comparison with different methods. Best viewed on Screen. To aid comprehension, Chinese text lines in the image are translated into English and annotated using corresponding colors.}
  \label{fig:exp_main}
    \vspace{-0.6cm}
\end{figure*} 

\noindent \textbf{Subject Fidelity Feedback Learning.}
The foreground extension detector $S_{\theta}$, after the offline training, is used as a reward model to supervise PosterMaker to improve subject fidelity. Specifically, assuming the reverse process has a total of $T'$ steps, we follow ReFL~\cite{ReFL} to first sample $z_{T'} \sim \mathcal{N}(0, 1)$ and after $T'-t'$ steps of inference ($z_{T'} \rightarrow z_{T'-1} \rightarrow \cdots \rightarrow z_{t'}$), we obtain $z_{t'}$, where $t' \sim [1, t_1]$. Then, we directly perform a one-step inference $z_{t'} \rightarrow z_{0}$ to accelerate the reverse process. Furthermore, $z_{0}$ is decoded to the generated image $x_{0}$. The detector $S_\theta$ predicts the foreground extension score for $x_{0}$, and this score is used as the reward loss to optimize the generator $G_\phi$ (i.e., PostMaker). The reward loss is defined as follows:
\vspace{-0.2cm}
\begin{equation}
\begin{split}
\mathcal{L}_{\text{reward}}(\phi) = -\mathbb{E}&_{(x, c, m) \sim \mathcal{D}_{\text{train}}, t' \sim [1, t_1],  z_{T'} \sim \mathcal{N}(0, 1)} \\
& \log \sigma \left( 1-  S_{\theta}( G_{\phi}(z_{T'},x,  c, m, t'), m) \right),
\end{split}
\end{equation}
where $x, c, m$ sampled from the train data $\mathcal{D}_{\text{train}}$, represent the subject image, control conditions, and subject mask respectively. To avoid overfitting, we don't calculate reward loss for the cases where the foreground extension score is below 0.3. Our total training loss is defined as:
\vspace{-0.2cm}
\begin{equation}
  \mathcal{L}_{\text{total}} = \mathcal{L}_{\text{denoise}}+\lambda\mathcal{L}_{\text{reward}},
  \vspace{-0.2cm}
\end{equation}
where $\lambda$ is the hyperparameter to adjust the weight of reward loss and the denoise loss.

\vspace{-0.1cm}
\subsection{Training Strategy} \label{sec:multi_stage_training}
\vspace{-0.1cm}

To efficiently train PosterMaker, this paper introduces a two-stage training strategy, as shown in~\cref{fig:multi_stage_training}, aimed at decoupling the learning for text rendering and background image generation. Specifically, in the first stage, the training task is local text editing. We fronze SceneGenNet and only the TextRenderNet and adapter are optimized. Since we initialize SceneGenNet with pre-trained weights of inpainting-controlnet~\cite{sd3-inpainting}, it can fill the local background well thus TextRenderNet can focus on learning text generation. In the second stage, the training task is subject-based text-to-image generation. Here we froze TextRenderNet and only train the SceneGenNet. In this stage, SceneGenNet focuses on learning poster scenes and creative design from the train data. 
Notably, Stage 1 learns local text editing/inpainting and Stage 2 learns background inpainting, thus the input images indicating the area to inpaint are different (See \cref{fig:multi_stage_training}). 
With such a two-stage training strategy, TextRenderNet and SceneGenNet can be efficiently optimized since they can focus on their specific tasks. 

%% file: sec/4_experiments.tex
 \vspace{-0.2cm}
\section{Experiments}
\vspace{-0.1cm}
\subsection{Experimental Setup}
\vspace{-0.1cm}
\noindent \textbf{Dataset.}
We crawl product posters from online e-commerce platforms to construct our training set. Our training data mainly consists of Chinese posters, we first employ PPOCRv4 model~\cite{ppocrv4} to extract the text content and their bounding boxes from the images as a coarse annotation. And we ask some annotators to further refine the bounding boxes and correct the text content to improve the annotation quality. Resulting in a dataset containing 160k images. We generate image captions with GPT4-o~\cite{gpt4o} and extract foreground subject masks with U$^2$-Net~\cite{U2Net} and VitMatte~\cite{yao2024vitmatte}. We randomly select 302 images for evaluation and leave the rest for training. To better evaluate the performance of our method, we use LLM~\cite{internlmxcomposer2} to generate some background prompts and text layouts as evaluation samples, after manually checking and removing those irrational ones, we obtain another 198 evaluation samples to form a final evaluation set named PosterBenchmark containing 500 samples. 

\begin{figure*}[!h]
  \centering
\includegraphics[width=\textwidth]{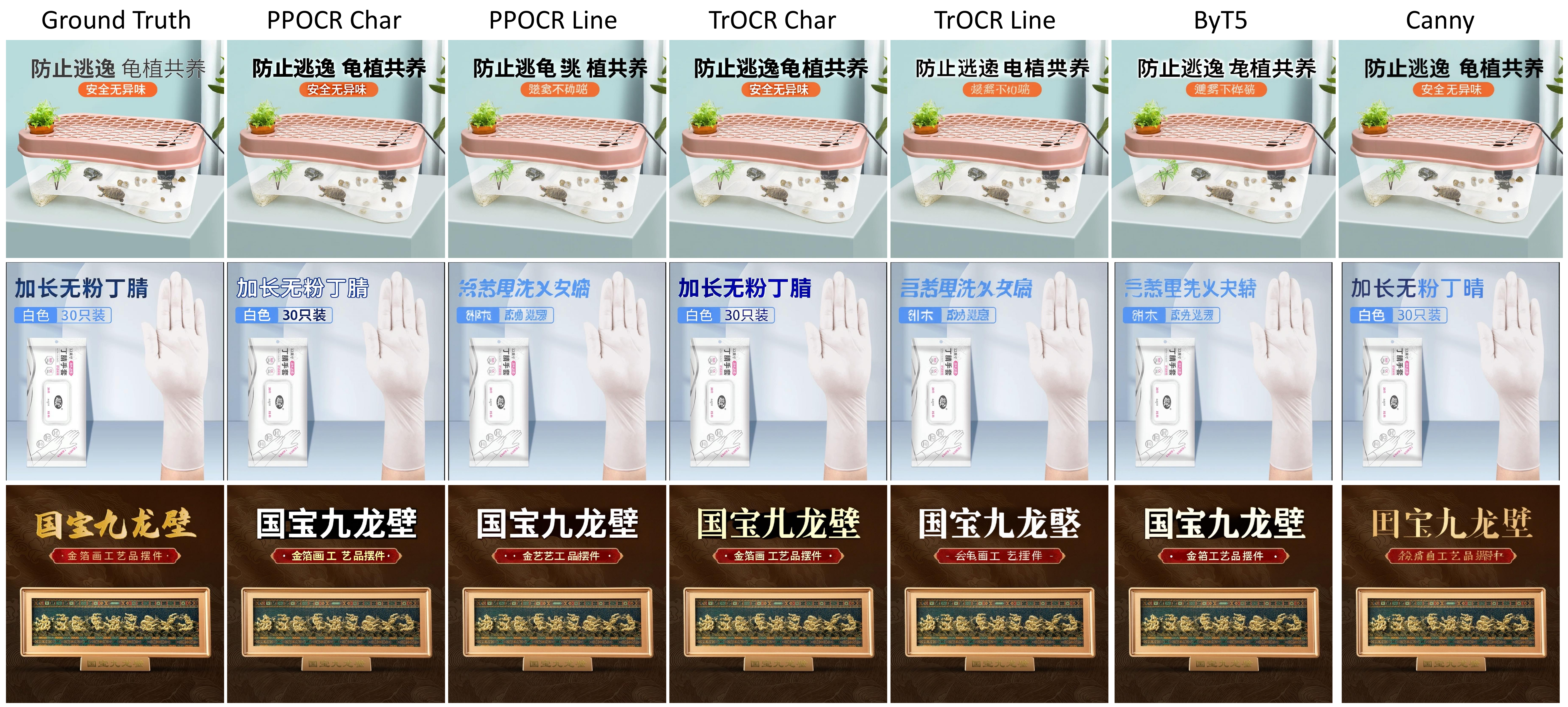}
 \vspace{-0.8cm}
  \caption{Qualitative comparison using various text features. It is obvious that the character-level OCR features we used (PPOCR Char) are the most effective at maintaining character accuracy.}
  \label{fig:exp_feat_ablation}
    \vspace{-0.5cm}
\end{figure*}

\noindent \textbf{Evaluation Metrics.}
We follow Anytext~\cite{tuo2023anytext} to evaluate text rendering accuracy using two metrics: sentence accuracy (Sen. Acc) and normalized edit distance (NED). Specifically, we crop the text line from the generated image according to the provided bounding box and utilize the OCR model~\cite{duguangocr} to predict the content $s_{\text{pred}}$ of the generated text line. We denote the ground truth text content as $s_{\text{gt}}$. A text line is considered to be correctly generated if $s_{\text{pred}} = s_{\text{gt}}$; this condition is used to calculate Sen. Acc. Additionally, we compute the normalized edit distance (NED) between $s_{\text{pred}}$ and $s_{\text{gt}}$ to measure their similarity. We further calculate FID~\cite{DBLP:conf/nips/HeuselRUNH17} to measure the visual quality and CLIP-T~\cite{ruiz2023dreambooth} metric for evaluating text-image alignment.

\noindent \textbf{Implementation Details.}
 Our SceneGenNet is initialized from pre-trained SD3 Inpainting-Controlnet~\cite{sd3-inpainting} and TextRenderNet is initialized from SD3~\cite{esser2024scaling} weight with the same configuration as in ~\cite{sd3-softedge}.
For Subject Fidelity Feedback Learning, we follow existing work~\cite{ReFL} to uniformly sample $t'$ between $[1,10]$. Within this range, the one-step inference result of image $x_0$ from $t'$ is close to the full inference result.
 The weight coefficient of $\lambda$ is set to 0.0005. The learning rate is set to 1e-4 and the batch size is set to 192. We train our framework for 26k and 29.5k steps for training stage1 and stage2, respectively. 
Finally, PosterMaker was trained on 32 A100 GPUs for 3 days.
 During the sampling process, based on the statistical information, a maximum of 7 lines of text and 16 characters per line of text are selected from each image to render onto the image, as this setting can cover most situations in the dataset.

\vspace{-0.1cm}
\subsection{Comparison with Prior Works}
\vspace{-0.1cm}
\noindent \textbf{Baseline methods.} We carefully designed the following baseline approaches\footnote{Details can be found in the Appendix.} based on existing open-sourced techniques for comparative analysis. 
\textbf{SD3\_inpaint\_byt5}: We encode the text content into prompt embeddings using ByT5~\cite{xue-etal-2022-byt5} and employ an adapter to map these embeddings to the original prompt embedding space of SD3 before feeding them into the controlnet, which enables the controlnet to render multilingual text.
\textbf{SD3\_canny\&inpaint}: First render the text into a white-background image and extract the canny edge from it as control. Then finetune a pre-trained SD3 canny controlnet together with an inpainting controlnet to achieve multilingual text rendering.
\textbf{Anytext}: It is the SOTA open-sourced T2I method that supports multilingual text rendering and its text editing mode supports text inpainting~\cite{tuo2023anytext}. So we directly finetune it on our data using its text editing training pipeline.
\textbf{SD3\_inpaint\_Anytext}: First generate the background with SD3 inpainting controlnet, then render the text on the corresponding region using Anytext.
\textbf{Glyph-ByT5-v2} and \textbf{GlyphDraw2}: They are both the SOTA T2I methods that support multilingual text rendering~\cite{liu2024glyph-v2, ma2024glyphdraw2}. However, they don't have open-sourced pre-trained weights, so we reproduced them on our dataset. And we added an inpainting controlnet for them to support subject-preserved generation.

\begin{table}
    \begin{center}
        \scalebox{0.7}{
            \begin{tabular}{ 
            l  @{\hskip2pt}|c@{\hskip4pt}c@{\hskip4pt}c@{\hskip4pt}c@{\hskip4pt}c@{\hskip1pt}} 
               Model & Sen. ACC $\uparrow$ &  NED $\uparrow$ & FID $\downarrow$ & CLIP-T $\uparrow$  &FG Ext. Ratio $\downarrow$\\
            \hline    
            SD3\_inpaint\_AnyText& 52.78\%& 75.27\%& 100.87& 26.90 &14.82\%\\
            SD3\_inpaint\_byt5& 52.28\%& 86.57\%& 65.45& 26.71 &14.60\%\\
            AnyText&  63.90\%& 82.81\%& 71.27& 26.69 & 19.25\%\\
            Glyph-ByT5-v2 & 69.54\% & 87.65\% &  79.23&  26.60 & 18.91\%\\
            SD3\_canny\&inpaint& 80.75\%& 92.75\%& 67.19& 27.03 & 14.38\%\\
            GlyphDraw2 & 86.14\% & 96.78\% & 72.49& 26.72 & 16.52\%\\ 
        
            \hline
            GT (w/ SD1.5 Rec.)& 76.95\%& 89.91\%& - & - & - \\
            GT (w/ SD3 Rec.)& 98.09\%& 99.36\%& - & - & - \\ 
            GT& 98.53\%& 99.59\%& -& - & - \\
            
            \hline
            Ours (SD1.5)& 72.12\%& 88.01\%& 68.17& 26.93 &-\\
            
            Ours  & \textbf{93.36\%}& \textbf{98.39\%}& \textbf{65.35}& \textbf{27.04} &\textbf{11.57}\%\\
        
            \end{tabular}
        }  
        \vspace{-0.3cm}
        \caption{Comparison with baseline methods.}
        \label{table:baseline_cmp}
    \end{center}
    \vspace{-1.1cm}
\end{table}

\noindent \textbf{Quantitative Comparison.}
We trained all baseline models on the same dataset, and then quantitatively compared all methods on the PosterBenchmark, as shown in~\cref{table:baseline_cmp}. It is worth noting that SD3 is used as the base model by default, but since we observed that the SD1.5 VAE leads to significant error in reconstruction, to enable a more equitable comparison between our method and AnyText (SD1.5 architecture), we also implemented an SD1.5 version of PosterMaker with the same experimental setup as AnyText.
As the VAEs, especially SD1.5, introduce some reconstruction error and the OCR model may incorrectly recognize some characters, we also report the metrics on ground truth images as an upper bound.
As shown in ~\cref{table:baseline_cmp}, our method achieves the best performance on all metrics. Notably, on text rendering metrics Sen. ACC and NED, our model outperforms the baselines by an impressive margin and is already close to the upper bound. The promising results demonstrate the effectiveness of the proposed PosterMaker.

\noindent \textbf{Qualitative Comparison.}
The results are shown in \cref{fig:exp_main}. Compared to the baselines, our PosterMaker generates more readable and accurate poster images with texts, particularly for smaller texts. Notably, as an end-to-end generation method, PosterMaker automatically creates underlays to enhance the contrast between text and background, effectively highlighting the text. This feature is crucial in product poster design for capturing viewers' attention. These findings demonstrate that our PosterMaker successfully learns the distribution of posters created by human designers.

\vspace{-0.1cm}
\subsection{Ablation Study and Analysis}
\vspace{-0.1cm}
\begin{table}

    \begin{center}
        \scalebox{0.8}{
        \begin{tabular}{ 
        l  @{\hskip3pt}|c|c@{\hskip3pt}c@{\hskip3pt}} 
           Text Feature&Type & Sen. ACC&  NED\\
        \hline    
        ByT5& textual feat. & 33.48\%& 54.50\%\\
        Canny &img & 81.50\%& 92.72\%\\

        TrOCR Line& visual feat.& 26.58\%& 49.46\%\\

        TrOCR Char  &visual feat.& 94.27\%& 98.54\%\\ 
        
        PPOCR Line&visual feat.& 38.91\%&53.86\%\\

        PPOCR Char (Ours)&visual feat.&  \textbf{95.15\%}& \textbf{98.75\%}\\
        
        \hline
        GT (w/o Rec.)  &-& 98.53\%& 99.59\%\\
 GT (w/ SD3 Rec.)& - & 98.09\% &99.36\%\\
        \end{tabular}
    }  
        \vspace{-0.3cm}
	    \caption{Quantitative comparison using various text features.}
        \vspace{-0.1cm}
	    \label{table:text_feature}
	\end{center}
\end{table}

\begin{table}
    \vspace{-0.5cm}
    \begin{center}
        \scalebox{0.7}{
        \begin{tabular}{ 
        l  @{\hskip3pt}|c@{\hskip3pt}|c@{\hskip3pt}|c@{\hskip3pt}|c@{\hskip3pt}|c@{\hskip3pt}} 
           Method& FG Ext. Ratio$\downarrow$ & Sen. ACC $\uparrow$ &  NED$\uparrow$ & FID$\downarrow$ & CLIP-T$\uparrow$ \\
        \hline    
        Ours & \textbf{11.57\%} & \textbf{93.36\%} & \textbf{98.39\%} & 65.35 & \textbf{27.04} \\
        Ours w/o $\mathcal{L}_{reward}$ & 15.05\% & 93.11\%& 98.21\% & \textbf{65.10}& \textbf{27.04}\\
        \end{tabular}
    }  
    \vspace{-0.3cm}
	    \caption{Evaluation on the subject fidelity feedback learning.}
     \label{table:exp_reward_loss}
	\end{center}
    \vspace{-1cm}
\end{table}

\noindent \textbf{How to achieve high text rendering accuracy?}
We conduct experiments to explore the effectiveness of different control conditions for visual text rendering. Due to the fact that text rendering accuracy is primarily determined by the first training stage,  we discard the second training stage in this experiment to save computational resources. The results are summarized in \cref{table:text_feature}. We observed several valuable experimental results: 1) The use of char-level features significantly outperforms previous line-level features, benefiting from finer-grained representation. This explains why previous methods~\cite{tuo2023anytext, chen2024diffute, ma2024glyphdraw2},  achieve inferior performance (PPOCR Line is used in~\cite{tuo2023anytext, ma2024glyphdraw2}, TrOCR Line is used in ~\cite{chen2024diffute}). 
Recent concurrent works~\cite{wang2024textmastero, ma2024chargen} have also found similar experimental findings as ours.
2) Char-level feature representation is superior to low-level image features such as Canny.
3) PPOCR outperforms TrOCR, which is attributed to PPOCR being a multi-language OCR model, while TrOCR is an English version model. 4) Even though TrOCR has not been trained on multi-language text data, it still achieves decent results, likely because it extracts universal visual structural features. 5) ByT5 extracts char-level features but the performance is inferior to OCR features, because it extracts semantic features rather than character structural features, while T2I models' text rendering capability relies more on character structural features. 
We present visualization results in \cref{fig:exp_feat_ablation}. We observe that when using line-level features as a control, the generated text occasionally becomes completely unrecognizable. This suggests that line-level features are insufficient for achieving precise text rendering. Additionally, it is evident that using canny control always introduces stroke artifacts, particularly in smaller texts (as seen in row 3 of \cref{fig:exp_feat_ablation}). This further demonstrates that canny control is also not an ideal condition for text rendering.
In summary, the char-level feature extracted by PPOCR performs optimally and the accuracy is already close to the upper bound, indicating \textit{the discriminative char-level visual feature is the key to achieve high text rendering accuracy}. 

\begin{figure}[t!]
    \centering
    \includegraphics[width=1.0\linewidth]{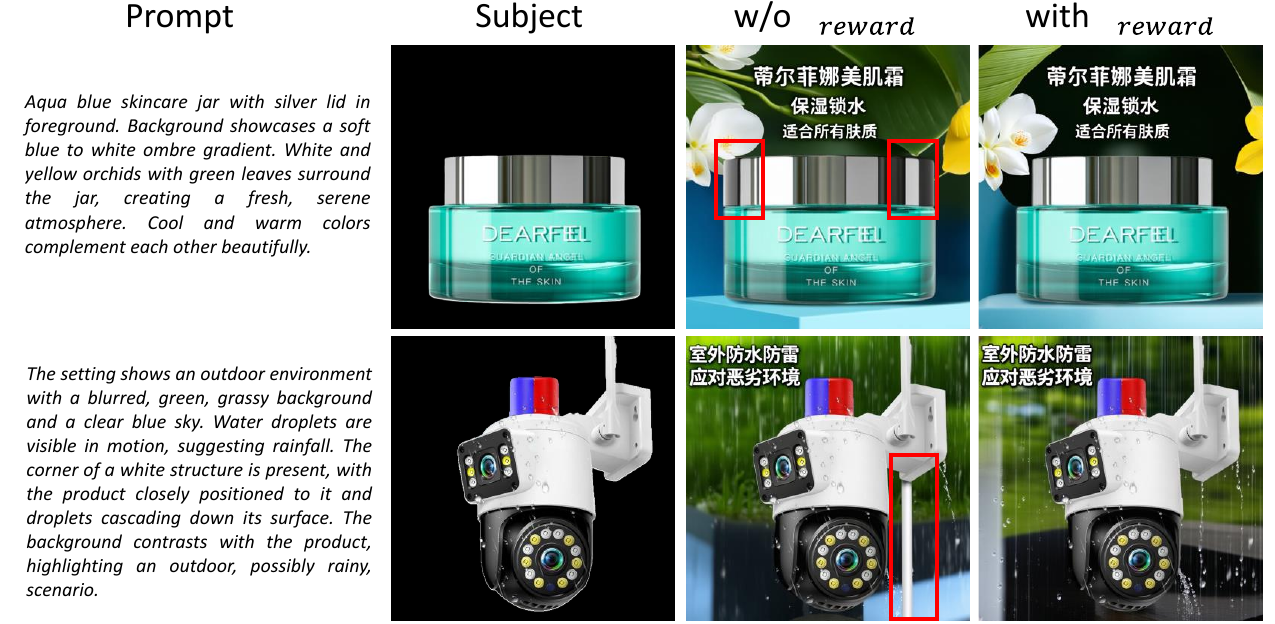}
    \vspace{-0.7cm}
    \caption{Visual examples showing the effect of $\mathcal{L}_{reward}$.}
    \label{fig:exp_reward_model}
    \vspace{-0.5cm}
\end{figure}

\noindent \textbf{Effectiveness of subject fidelity feedback learning.}
We calculate the foreground extension ratio (termed as FG Ext. Ratio) by asking human annotators to manually check each generated image whether the foreground subject is incorrectly extended. As demonstrated in ~\cref{table:exp_reward_loss}, training our model with $\mathcal{L}_{reward}$ effectively reduces FG Ext. Ratio by 3.4\%, while maintaining subtle variations in other performance metrics. Representative visual examples are presented in ~\cref{fig:exp_reward_model}.  Besides, our model outperforms baseline methods in FG Ext. Ratio (see ~\cref{table:baseline_cmp}).
 These results show the efficacy of our proposed subject fidelity feedback learning approach in mitigating foreground extension artifacts.

%% file: sec/6_conclusion.tex
\vspace{-0.2cm}
\section{Conclusion}
\vspace{-0.1cm}
The application of image generation in poster creation is often impeded by subpar text rendering and inconsistent subjects. To address these challenges, this paper introduces a novel framework, PosterMaker, which synthesizes aesthetically pleasing product posters with accurate and harmonious texts and contents. Moreover, we reveal that the key underlying successful multilingual text rendering is the construction of robust character-level visual text representations.  Additionally, we propose subject fidelity feedback learning to mitigate inconsistencies in subjects. Through extensive experiments, our method demonstrates a significant improvement in both high-precision text generation and subject fidelity. These findings not only advance poster generation but also inspire future research on T2I models.

%% file: sec/acknowledgments.tex
\section*{Acknowledgments}
This work was supported by the National Nature Science Foundation of China (62425114, 62121002, U23B2028, 62232006, 62272436) and Alibaba Group (Alibaba Research Intern Program).

%% file: sec/X_suppl.tex
\clearpage
\setcounter{page}{1}
\maketitlesupplementary

Due to space limitations, we were unable to present all experimental results in the main text. In this supplementary material, we will give more details about our experiments and present additional results.

\section{Implementation Details}
\noindent \textbf{Training and Inference.}
We fully follow the settings of SD3~\cite{esser2024scaling}. During training, the denoise loss $\mathcal{L}_{\text{denoise}}$ uses simplified flow matching, also known as 0-rectified flow matching loss \cite{liu2022flow}. In inference, we also use the inference method of flow matching, with 28 inference steps. 

\noindent \textbf{TextRenderNet and SceneGenNet.}
TextRenderNet and SceneGenNet have an architecture similar to SD3~\cite{esser2024scaling}, composed of multiple MM-DiT Blocks. In our implementation, TextRenderNet consists of 12 layers of MM-DiT Blocks, while SceneGenNet consists of 23 layers of MM-DiT Blocks. 
The output of the $N_i$-th block of SceneGenNet is first added with the output of the $\left\lceil \frac{N_i}{2} \right\rceil$-th block of TextRenderNet, and then add to the $N_i$-th SD3 block.

\noindent \textbf{Classifier-Free Guidance.}
We use CFG during inference, with a CFG scale of 5. Additionally, since the ``prompt'' inputted to TextRenderNet is not a caption but a text representation, the negative one for CFG is set to a zero vector. During training, we randomly drop the text representation to a zero vector with 10\% probability.

\noindent \textbf{The Setting of  $t_1$ in Reward Loss. }
We follow~\cite{ReFL} to train the reward loss at the last 10 inference steps, i.e., we set $t_1$ to 10. Within the range of $t '\sim [1, t_1] $,  the result of the image $x_0$ obtained by one-step inference is close to the result of complete inference.

\noindent \textbf{Details about Metric Calculation.} Our evaluation benchmark contains samples generated by LLM~\cite{internlmxcomposer2} thus there is no ground truth for these samples. Therefore, we exclude these LLM-generated samples when calculating metrics that depend on ground truth images, i.e., FID metric for all experiments, text accuracy metrics for GT (with and without VAE reconstruction) and results for ablation on different text features.

\noindent \textbf{About ground truth for training Foreground Extension Detector.}
We treat the task of detecting foreground extension as a binary classification problem and ask annotators to manually label the ground truth.

\begin{figure}[t]
    \centering
    \includegraphics[width=0.5\textwidth]{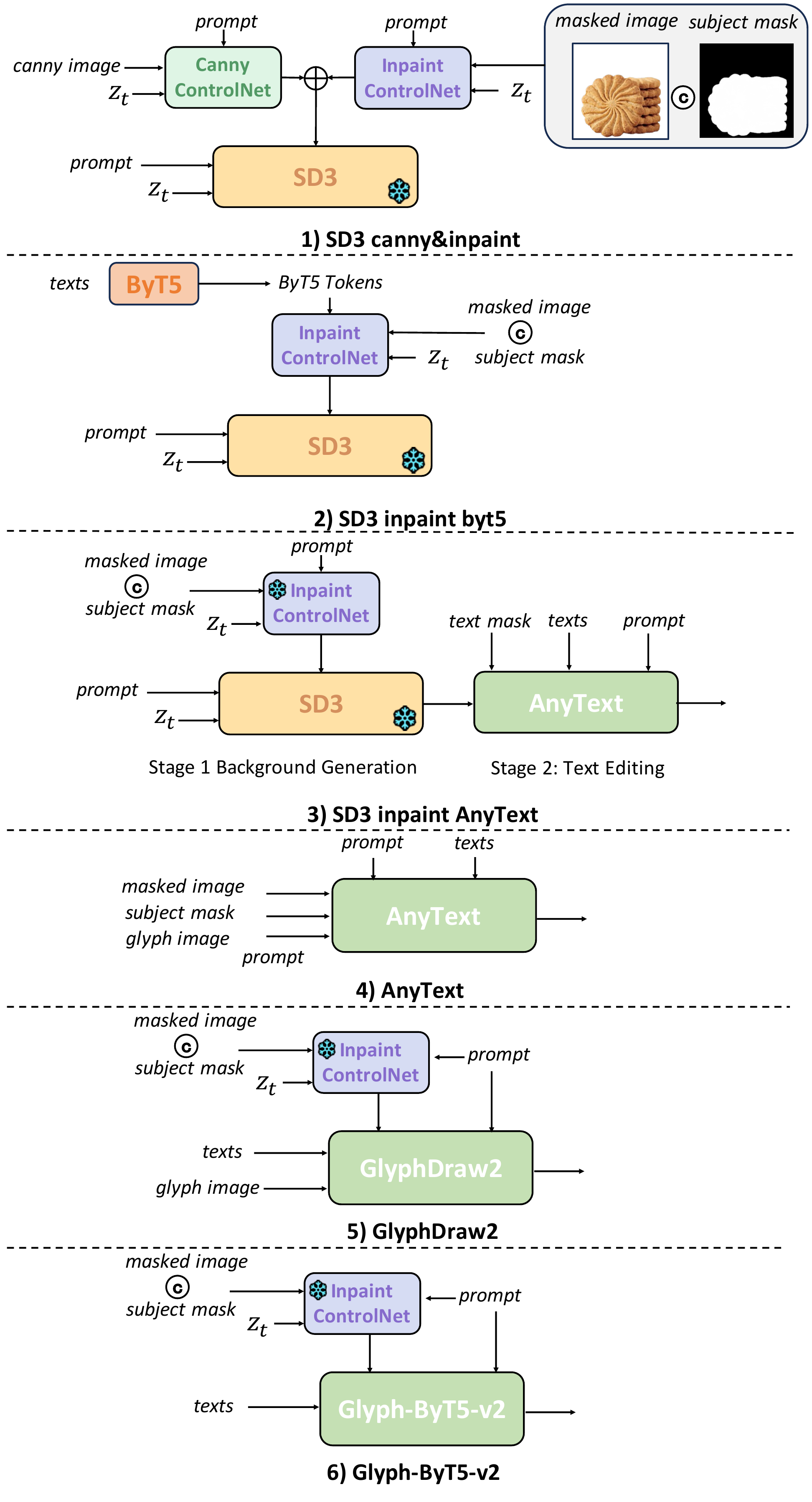}
    \caption{Detailed illustration of the implementation of the different baseline methods.}
    \label{fig:baseline_details}
\end{figure}

\section{Baseline Details}
We carefully designed 6 baseline approaches based on existing techniques for comparative analysis. The details are shown in~\cref{fig:baseline_details}. For 1) SD3\_inpaint\_byt5, 2) SD3\_canny\&inpaint, and 4) AnyText, we fine-tune them on our 160K dataset for the poster generation task. Meanwhile, 3) SD3\_inpaint\_Anytext is a two-stage inference method. In the first stage, the pre-trained Inpaint ControlNet generates the background, and in the second stage, AnyText performs the text editing task, with AnyText also fine-tuned on the 160K dataset specifically for the text editing task. 
The Inpainting ControlNet is initialized from pre-trained SD3 Inpainting-Controlnet~\cite{sd3-inpainting} and Canny ControlNet is initialized from~\cite{sd3-softedge}. 
For 5) GlyphDraw2~\cite{ma2024glyphdraw2} and 6) Glyph-ByT5-v2~\cite{liu2024glyph-v2} are both the SOTA T2I methods that support multilingual text rendering. However, they neither have open-source pre-trained weights nor support subject input, so we reproduced them on our dataset by adding the pre-trained inpainting controlnet~\cite{sdxl-inpainting} to support the subject input.

\section{Scalable Training for Text Rendering}
Our proposed two-stage training strategy allows the model to learn two different capabilities (i.e., text rendering and scene generation) separately, enabling more flexibility with distinct datasets for each phase. Recent text rendering methods~\cite{tuo2023anytext, chen2024diffute, liu2024glyph, liu2024glyph-v2} typically train their models on datasets containing millions of samples. To verify the potential of further improving our performance with more training data, we build a large dataset with 1 million samples and we directly obtain the text annotations with PPOCRv4~\cite{ppocrv4} without manually annotating. And we use this dataset for the first stage of text rendering training and use the same 160k data for the second stage of scene generation learning. Compared to using 160k data in both of the previous stages, the text sentence accuracy significantly improved by 4.48\% (as shown in~\cref{table:scaling_training}), demonstrating that the multi-stage training strategy is flexible and scalable. However, in the main experiments, we select to report the performance of our model training only on 160k data for fair comparison with the baselines.

\begin{table}[h]
    \begin{center}
        \scalebox{0.9}{
        \begin{tabular}{ 
        l@{\hskip9pt}|c@{\hskip9pt}c@{\hskip7pt}} 
        Data Size (St.1 \& St.2)  & Sen. ACC&  NED\\
        \hline    
        160k \& 160k  & 93.11\% & 98.21\% \\

        1M \& 160k &  \textbf{97.59\%} & \textbf{99.38\%}

        \end{tabular}
     }
	    \caption{Quantitative comparison with different data sizes for text rendering training.}
	    \label{table:scaling_training}
	\end{center}
        \vspace{-1.0cm}
\end{table}

\section{Discussion on advantages of end-to-end over two-stage methods.} 
\begin{figure}[h]
    \vspace{-0.3cm}
    \centering
    \includegraphics[width=1.0\linewidth]{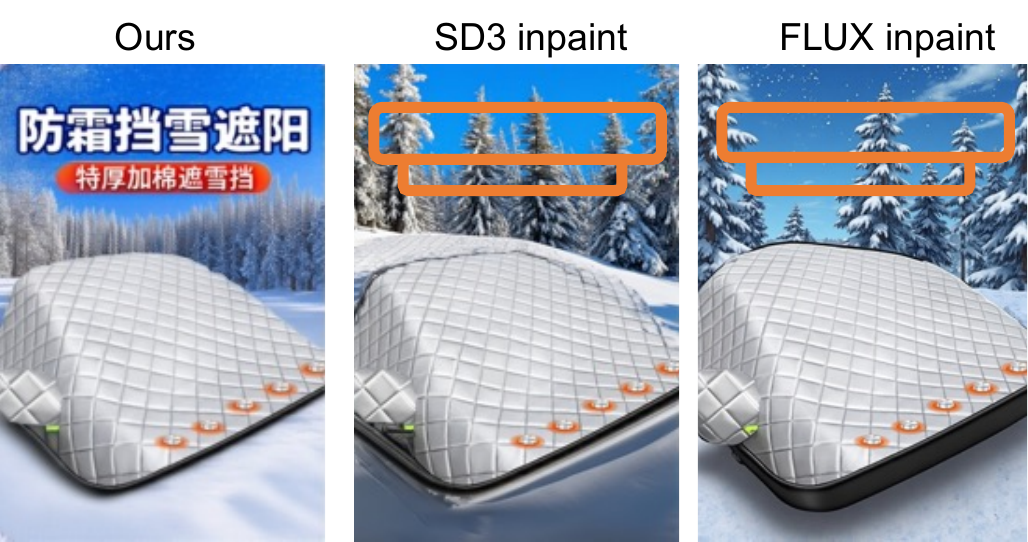}
    \vspace{-0.4cm}
    \caption{Showcases for end-to-end and two-stage methods.}
    \label{fig:R1}
\end{figure}

The main weakness of two-stage methods (first inpaint background, then render text) is their inability to consistently provide a clean background for texts (see \cref{fig:R1}, reducing text readability, especially with complex backgrounds. In contrast, one-stage methods generate texts and backgrounds simultaneously, enabling them to create a clean backdrop or underlays that enhance text visibility.

\vspace{-0.2cm}

\section{Text Position Control}
\label{sec:text_loc_control}
\begin{figure}[h]
    \centering
    \includegraphics[width=0.5\textwidth]{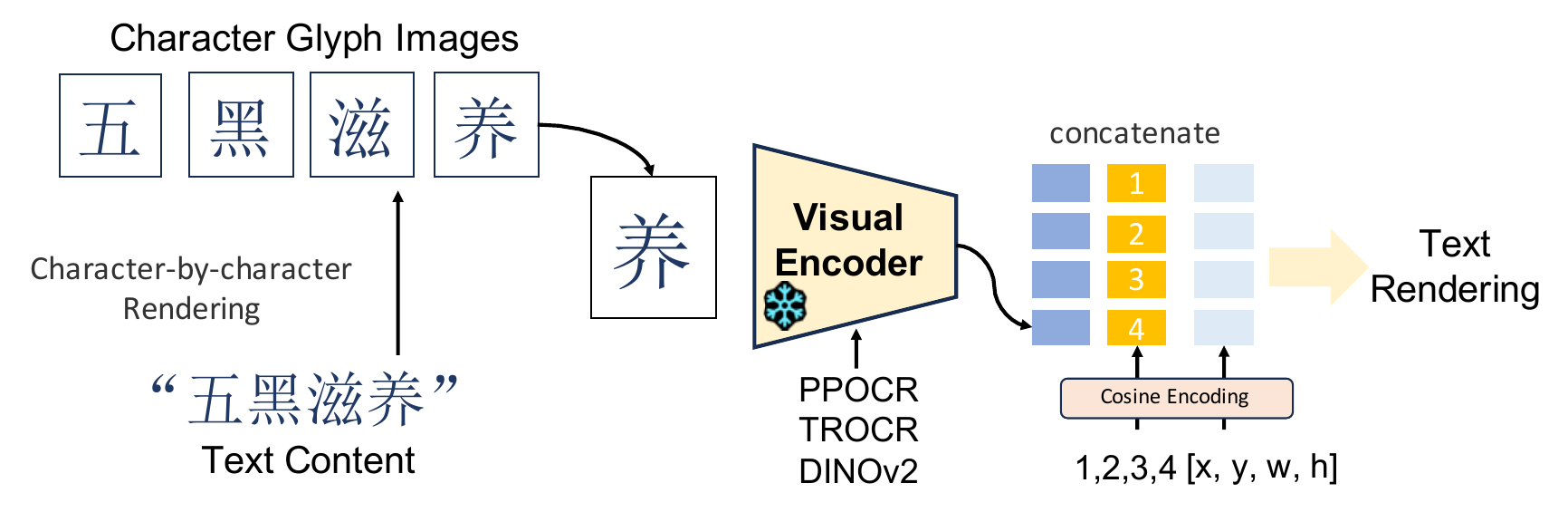}
    \caption{Detailed illustration of how we construct the position embedding for controlling the text position.}
    \label{fig:text_pos_control}
    
\end{figure}

The position control of PosterMaker uses a very straightforward approach (as shown in~\cref{fig:text_pos_control}), mapping the text bounding box to cosine position encoding, which is then concatenated with text features and used as the input to TextRenderNet. To demonstrate our method's effectiveness, we evaluate the bounding box IoU (Intersection of Union) metric as follows: 1) we employ OCR model to extract texts from the generated image. 2) For each ground truth text, we identify the best-matched OCR-detected text based on edit distance and then calculate the IoU between their corresponding bounding boxes. We average the IoU score over all the samples to obtain mean IoU (termed mIoU). And we also report IoU@R which indicates the proportion of samples with IoU higher than $R$. As shown in~\cref{table:text_iou}, our method achieves a high mIoU of 84.65\% and 93.94\% samples have an IoU score higher than 0.7. These promising results prove that our text position control method is simple yet effective.

\begin{table}
    \begin{center}
        \scalebox{0.9}{
        \begin{tabular}{ 
        l  @{\hskip7pt}|c@{\hskip7pt}|c@{\hskip7pt}|c@{\hskip7pt}} 
           Method& mIoU & IoU@0.5 & IoU@0.7 \\
        \hline    
        Ours & 84.65\% & 97.18\% & 93.94\% \\
        \end{tabular}
    }  
	    \caption{Evaluation on text location accuracy.}
     \label{table:text_iou}
	\end{center}
    \vspace{-0.4cm}
\end{table}

\section{Comparison Between GlyphByT5 and PosterMaker}
GlyphByT5~\cite{liu2024glyph, liu2024glyph-v2} are recently proposed visual text rendering methods that achieve high text rendering accuracy. And we will discuss some differences and internal connections between our PosterMaker and GlyphByT5 on how to control text rendering.
\begin{itemize}
    \item Text position control: GlyphByT5 achieve text position control by modifying the original cross-attention module with their proposed region-wise multi-head cross-attention. In contrast, our PosterMaker encodes the text location directly into the character-level text representation to accomplish text position control. As discussed in~\cref{sec:text_loc_control}, our approach is both simple and effective for precise text location control.

    \item Text content control: both GlyphByT5 and our PosterMaker control the generation of text content by constructing suitable text representation. Specifically, in this work, we claim that the key to achieve accurate text rendering is to extract \textbf{character-level visual} features as the control condition and carefully construct a robust text representation based on off-the-shelf OCR model~\cite{ppocrv4}. In GlyphByT5, the authors also extract \textbf{character-level text} features, but with a \textit{textual} encoder named ByT5~\cite{xue-etal-2022-byt5}. Then they propose glyph-alignment pre-training to align these \textit{textual} features with pre-trained \textit{visual} encoders DINOv2~\cite{oquab2023dinov2}. Additionally, they employ box-level contrastive learning with complex augmentations and a hard-mining strategy to enhance \textit{character-level} discriminativeness. We hypothesize that the primary reason both our method and GlyphByT5 achieve high text rendering accuracy is our shared goal of constructing a robust \textbf{character-level visual} representation. In fact, the ability of GlyphByT5's character-level visual representation is distilled from the pre-trained \textit{visual} encoder DINOv2, rather than inherited from the  pre-trained \textit{ textual} encoder ByT5 itself.   In order to verify our hypothesis and insights, we adopt a more direct approach to directly replace the PPOCR encoder in PosterMaker with DINOv2. As shown in~\cref{table:compare_glyphbyt5}, simply extracting character-wise visual features with DINOv2 can also achieve precise text rendering. This result further verifies our claim: the key to precise text rendering is to extract \textbf{character-level visual} features as the control condition.

\end{itemize}

\begin{table}[h]
    \begin{center}
        \scalebox{0.9}{
        \begin{tabular}{ 
        l  @{\hskip9pt}|c|c@{\hskip9pt}c@{\hskip7pt}} 
           Text Feature&Type & Sen. ACC&  NED\\
        \hline    
        PPOCR Line&visual feat.& 38.91\%&53.86\%\\

        PPOCR Char &visual feat.&  95.15\%& 98.75\%\\
        \rowcolor{gray!30}   
         DINOv2 Line& visual feat.& 4.25\% & 20.59\% \\
        \rowcolor{gray!30}  
         DINOv2 Char& visual feat.& 94.92\%  & 98.66\%\\
        \hline
        GT (w/o Rec.)  &-& 98.53\%& 99.59\%\\
        GT (w/ SD3 Rec.)& - & 98.09\% &99.36\%\\
        \end{tabular}
    }  
	    \caption{Quantitative comparison using various text features.}
	    \label{table:compare_glyphbyt5}
	\end{center}
\end{table}

\section{Visualization of Training Samples}
\begin{figure*}[!h]
  \centering
\includegraphics[width=0.95\textwidth]{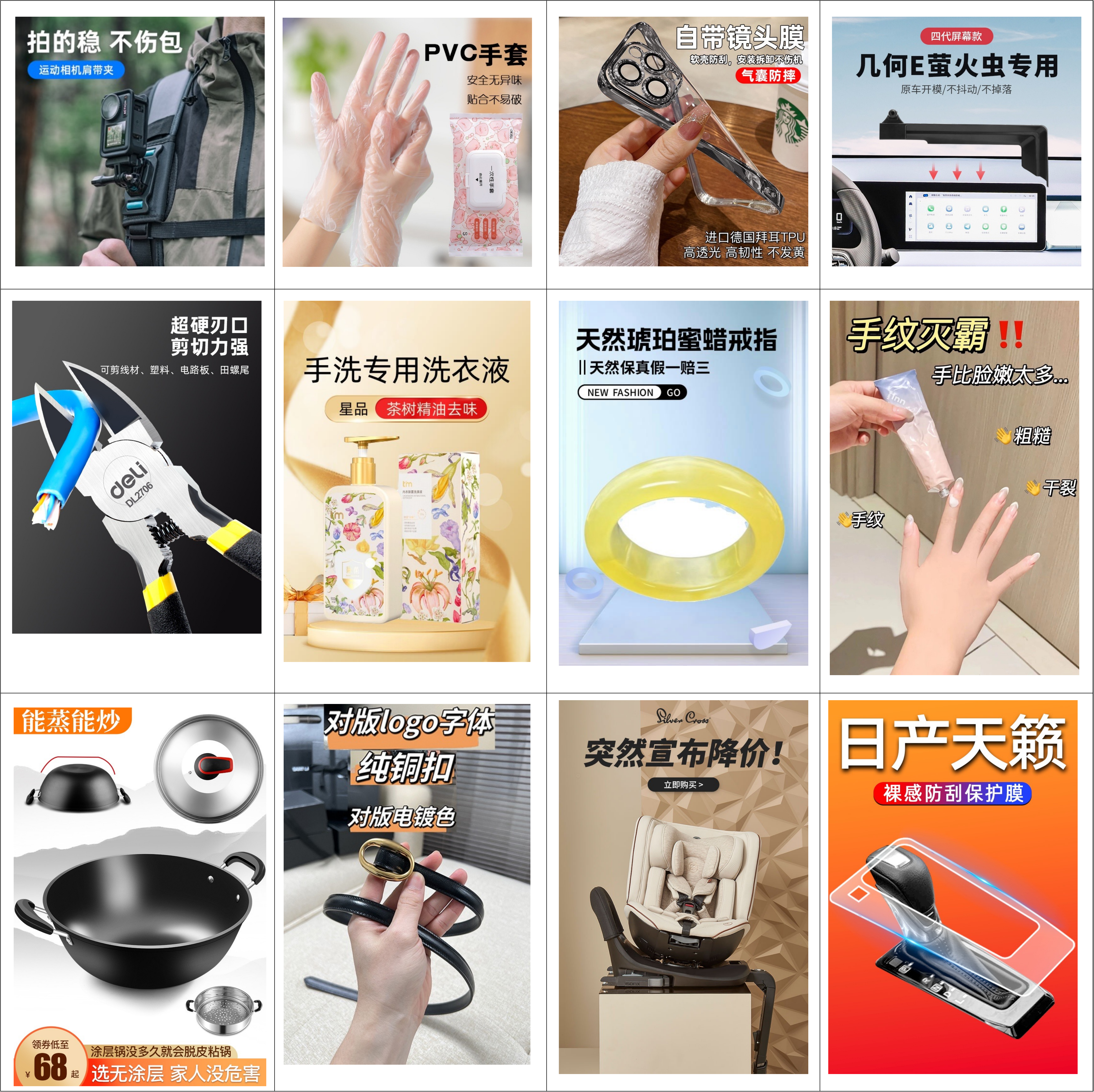}

  \caption{Visualization of ground truth for some samples in the dataset.}
  \label{fig:dataset_example}
\end{figure*}

We present example training images from our dataset in\cref{fig:dataset_example}. The dataset predominantly consists of Chinese text, with a small portion of English text. Additionally, it includes challenging cases with small-sized text elements.

\section{The Generalization of Text Representation.}
\begin{figure*}[!h]
    \centering
    \includegraphics[width=1.0\textwidth]{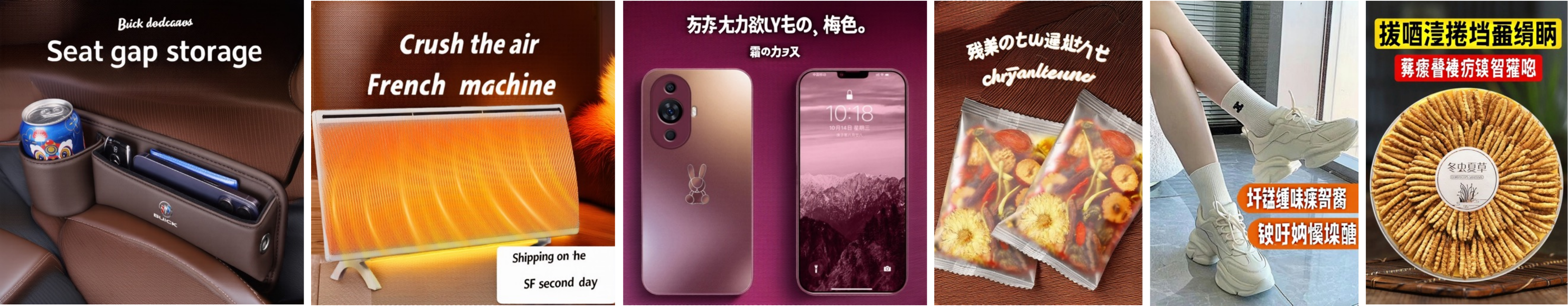}
    \vspace{-0.2cm}
    \caption{Visualization results on texts in English, Japanese, and uncommon Chinese characters.}
    \label{fig:multilingual_text_renderi}
\end{figure*}

PosterMaker is trained primarily on common Chinese data, with only a minimal amount of English data. Despite this, it demonstrates a notable level of generalization, enabling it to generate English, Japanese, and uncommon Chinese characters that were not included in the training set (as shown in~\cref{fig:multilingual_text_renderi}). In order to quantitatively evaluate the generalization capability of PosterMaker, we compared the accuracy of different text representations on uncommon characters using a randomly sampled uncommon character benchmark. The results show that our method can also generalize well to some characters that are unseen in the training set. Our performance is inferior to the canny baseline, likely because the canny baseline has been pre-trained on large-scale image data.

\begin{table}[h]
    \begin{center}
        \scalebox{0.9}{
        \begin{tabular}{ 
        l  @{\hskip9pt}|c|c@{\hskip9pt}c@{\hskip7pt}} 
           Text Feature&Type & Sen. ACC&  NED\\
        \hline    
        ByT5 & textual feat. & 2.01\%& 10.27\%\\
        Canny & img & \textbf{65.12\%}& \textbf{74.56\%}\\
        PPOCR Line& visual feat.& 8.34 \%& 15.84\%\\
        PPOCR Char & visual feat.&  61.54\%& 70.38\%\\
        \end{tabular}
    }  
	    \caption{Quantitative comparison of the rendering results of different text features on uncommon characters.}
	    \label{table:unseen_char}
	\end{center}
\end{table}

\begin{table}
    \begin{center}
        \scalebox{0.9}{
        \begin{tabular}{ 
        l  @{\hskip3pt}|c@{\hskip3pt}|c@{\hskip3pt}|c@{\hskip3pt}} 
           Method& Precision & Recall & F1 Score\\
        \hline    
        RFNet (our impl.) & 76.52\% & 75.52\% & 76.02\% \\
        RFNet (SAM) & 81.35\% & 80.99\% & 81.17\%\\
        Ours & \textbf{83.52\%} & \textbf{84.81\%} & \textbf{84.16\%} \\
        \end{tabular}
    }  
	    \caption{Evaluation on different architectures of foreground extension detector.}
     \label{table:exp_reward_architecture}
	\end{center}
\end{table}

\begin{figure}[t!]
    \centering
    \includegraphics[width=\linewidth]{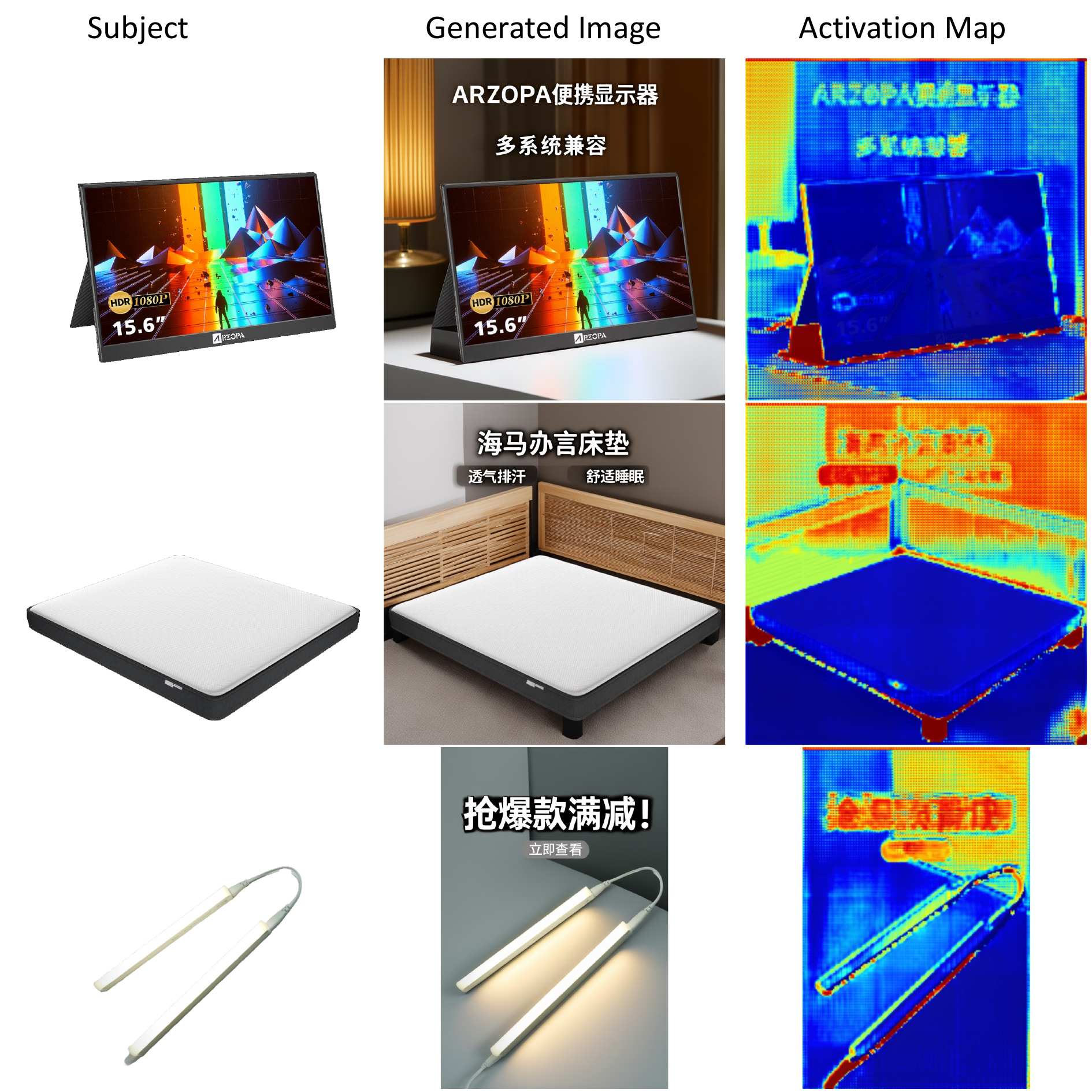}
    \caption{Class activation map of the foreground extension detector.}
    \label{fig:reward_model_attn}
\end{figure}

\section{Ablation about Foreground Extension Detector}
We collected 20k manually annotated images to train the foreground extension detector. We randomly selected 10\% samples as a validation set, while using the remaining 90\% for model training. We conduct ablation experiments on different architecture designs of the detector to verify the effectiveness of the proposed architecture. We implement 2 baselines: 1) \textbf{RFNet}~\cite{du2024towards}: we reimplemented RFNet based on the description in their paper~\cite{du2024towards}. Since we could not access their depth and saliency detection models, we modified our implementation to only use the product image and generated image as input, excluding the depth and saliency maps. 2) \textbf{RFNet(SAM) }: in this baseline, we replace the image encoder used in RFNet with the same SAM\cite{SAM} image encoder used in our method. As summarized in~\cref{table:exp_reward_architecture}, our proposed foreground extension detector outperforms the baselines by a considerable margin, which demonstrates its effectiveness.

In~\cref{fig:reward_model_attn}, we visualize the class activation map~\cite{zhou2016cvpr_cam} of our proposed foreground extension detector. As shown, we can observe a notably higher activation score in the extended foreground regions compared to other areas. This compelling evidence demonstrates that our detector has effectively learned to discern foreground extension cases, thereby it can serve as a robust reward model for fine-tuning PosterMaker to mitigate the foreground extension problem.

\section{Ablation about SceneGenNet}
SceneGenNet enables our model to perform background inpainting while preserve the subject so we cannot directly remove it. We replace it by SDEdit~\cite{meng2022sdedit} to achieve inpainting. As the results shown in~\cref{tab:Ablation_SceneGenNet}, replacing it results in a significant drop of performance.

\begin{table}[h]
\centering
\resizebox{0.48\textwidth}{!}
{
    \begin{tabular}{l|c|c|c|c}
    Model & Sen. ACC $\uparrow$ &  NED $\uparrow$ & FID $\downarrow$ & CLIP-T $\uparrow$ \\ \hline
    Ours w/o SceneGenNet &  90.53\%&  97.95\%& 79.44& 26.67\\ 
    Ours & \textbf{93.36\%}& \textbf{98.39\%}& \textbf{65.35}& \textbf{27.04} 
    \end{tabular}
}
\label{tab:Ablation_SceneGenNet}
\vspace{-0.2cm}
\caption{Comparison between SceneGenNet and SDEdit}

\end{table}

\section{Discussion on the impact of the test set size.}
To ensure a fairer comparison between PosterMaker and the baseline methods, we expanded the test set to 5,000 samples(10x the previous PosterBenchmark). The results are shown in~\cref{tab:comparison_big_test_set}, and the experimental conclusions remain consistent with the previous test set. Due to the calculation principle of the FID metric, increasing the test size leads to a significant decrease in the FID scores for all methods, but still maintains the same conclusion.

\begin{table}[!h]
\centering
\resizebox{0.48\textwidth}{!}
{
    \begin{tabular}{l|c|c|c|c}
    Model & Sen. ACC $\uparrow$ &  NED $\uparrow$ & FID $\downarrow$ & CLIP-T $\uparrow$ \\
    \hline
    Glyph-ByT5-v2 & 67.87\%& 86.23\%& 20.37&21.08\\ 
    SD3\_canny\&inpaint &  74.49\%&  88.78\%&  17.91&  20.79\\ 
    GlyphDraw2&  83.81\%&  96.49\%& 15.24& 20.67\\ \hline
    Ours &  \textbf{90.20\%}& \textbf{97.58\%}& \textbf{13.36}&   \textbf{21.36}\\ 
    \end{tabular}
}
\vspace{-0.2cm}
\caption{Comparison with baseline methods on 5,000 test samples.}
\label{tab:comparison_big_test_set}
\vspace{-0.2cm}
\end{table}

\section{Discussion on the meaningless texts generated outside target position.}
In our early experimental attempts about text rendering in poster generation, we found that the trained model sometimes generates meaningless texts outside the target area of the text, which will seriously affect the aesthetics. We conjecture that the main reason is that the ground truth images sometimes contain text outside the specified position. To solve this problem, we masked out the extra text during training and it solved most cases.

\begin{figure}[h]
    \centering
    \includegraphics[width=0.9\linewidth]{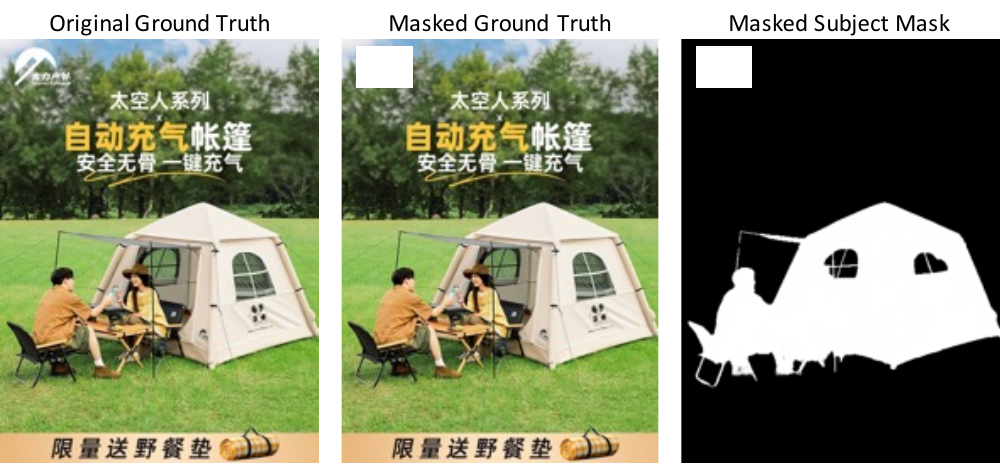}
    \caption{Example of our solution technique for meaningless texts and logos that generated outside target position. }
    \label{fig:discussion_meanfulless_text}
\end{figure}

Specifically, SceneGenNet is initialized from pre-trained SD3 Inpainting-Controlnet~\cite{sd3-inpainting}. In the second stage of training, we simultaneously mask out the regions of the untrained texts (usually those that are too small or just logos) both in the subject mask input to SceneGenNet and in the ground truth image used for loss calculation(as shown in~\cref{fig:discussion_meanfulless_text}). It is worth noting that although these small texts and logos are not included in the training, we have also annotated them to address the aforementioned issues. Finally, this technique makes the loss corresponding to the masked-out regions very close to zero so that the model will not learn these meaningless texts.

